\documentclass[10pt,twocolumn,letterpaper]{article}

\usepackage{iccv}
\usepackage{times}
\usepackage{epsfig}
\usepackage{graphicx}
\usepackage{amsmath}
\usepackage{amssymb}

\usepackage[pagebackref=true,breaklinks=true,letterpaper=true,colorlinks,bookmarks=false]{hyperref}

\usepackage{graphicx}
\usepackage{algorithm}
\usepackage{bm}
\usepackage{color}
\usepackage{multirow}
\usepackage{ifthen}
\usepackage{bbm}
\usepackage{pifont}
\usepackage{subcaption}
\newcommand{\cmark}{\ding{51}}%
\newcommand{\xmark}{\ding{55}}%
\graphicspath{{./figures/}}

\DeclareMathOperator{\cmax}{cmax}
\renewcommand{\paragraph}[1]{\vspace{0.1cm}\noindent\textbf{#1}\quad}

\newcommand{\method}[1]{\ifthenelse{\equal{#1}{full}}{Cross-sentence Relations Mining}{CRM}}

\newcommand{\best}[1]{{\textbf{\color{black}#1}}}
\newcommand{\second}[1]{#1}

\newcommand{\raymond}[1]{{\color{black}#1}}

\newcommand{\sgg}[1]{{\color{black}#1}}

\iccvfinalcopy 

\def\iccvPaperID{1623} 

\ificcvfinal\fi

\begin{document}

\title{Cross-Sentence Temporal and Semantic Relations in Video Activity Localisation}

\author{Jiabo Huang$^1$%
\thanks{Corresponding authors.}\\
{\tt\small jiabo.huang@qmul.ac.uk}
\and
Yang Liu$^2$%
\footnotemark[1]\\
{\tt\small yangliu@pku.edu.cn}
\and
Shaogang Gong$^1$\\
{\tt\small s.gong@qmul.ac.uk}
\and
Hailin Jin$^3$\\
{\tt\small hljin@adobe.com}
\and
$^1$Queen Mary University of London$\quad$
$^2$WICT, Peking University$\quad$
$^3$Adobe Research
}

\maketitle
\ificcvfinal\thispagestyle{empty}\fi

\begin{abstract}
Video activity localisation
has recently attained increasing attention
due to its practical values
in automatically localising the most salient visual segments 
corresponding to their language descriptions (sentences) from 
untrimmed and unstructured videos. 
For supervised model training,
a temporal annotation of both the start and end time index of each video segment for a sentence
(a video moment) must be given. This is not only very expensive but also sensitive to
ambiguity and subjective annotation bias, a much harder task than
image labelling.
In this work, we develop a more accurate weakly-supervised solution 
by introducing Cross-Sentence Relations Mining (CRM) in video moment
proposal generation and matching when only a paragraph description
of activities without per-sentence temporal annotation is available.
Specifically,
we explore two cross-sentence relational constraints:
(1) Temporal ordering and (2) semantic consistency among sentences in
a paragraph description of video activities. 
Existing weakly-supervised techniques only consider 	
within-sentence video segment correlations in training without 	
considering cross-sentence paragraph context. 
This can \sgg{mislead} due to
ambiguous expressions of \sgg{individual} sentences
with visually indiscriminate video moment proposals \sgg{in isolation}.
Experiments on two publicly available activity localisation datasets
show \sgg{the advantages} of our approach over the
state-of-the-art weakly supervised methods, 
especially so when the video
activity descriptions become more complex.
\end{abstract}


\section{Introduction}
\begin{figure}[ht]
\centering
\includegraphics[width=1.0\linewidth]{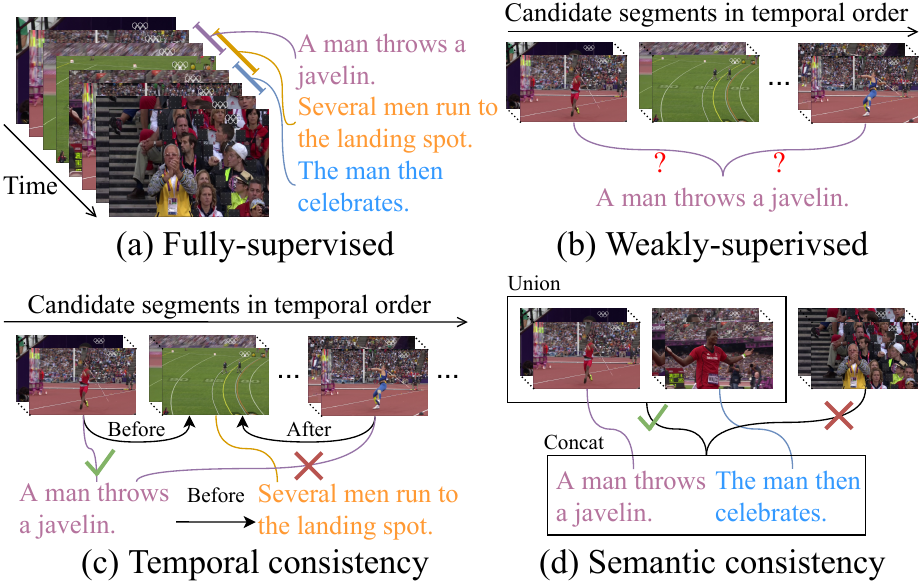}
\caption{
Different video activity localisation methods:
\textbf{(a)}
Given a paragraph description and the per-sentence temporal annotation
(start and end time index),
fully-supervised methods learn to
align sentences with ground-truth 
semantically matching video moments~\cite{chen2020hvtg,mun2020lgi}.
\textbf{(b)}
Without fine-grained temporal annotations,
weakly-supervised models often generate proposals of video segments
corresponding to sentences in a paragraph
before learning the best visual-text alignment~\cite{ma2020vlanet,lin2020scn}.
\textbf{(c)}
The \protect\method{abbr} model
explores the temporal order of different sentences in a paragraph
to minimise the ambiguities in matching the best video moments to
specific sentences in the context of a paragraph. 
\textbf{(d)}
To deal with ambiguous expressions in descriptions,
\protect\method{abbr} 
further explore plausible sentence expansion, 
\eg pairing two sentences (concatenation) as a more complex query 
to constrain the localisation of pairwise video moment proposals.
This explores cross-sentencing semantic consistency.
}
\label{fig:idea}
\vspace{-0.5cm}
\end{figure}
Video activity localisation by natural language 
is an important yet challenging task, 
which aims to localise temporally a video segment 
(moment\footnote{Video segment and moment are used interchangeably in this paper.}) that 
best corresponds to a query sentence
in an untrimmed (and often unstructured) video~\cite{mithun2019tga,duan2018wsdec}. 
Most of the existing methods address this task 
in a fully supervised manner~\cite{mun2020lgi,chen2020hvtg}, 
\ie the untrimmed video data are annotated by both
a paragraph description, 
in which each sentence is describing a video \textit{moment-of-interest} (MoI),
and per-sentence temporal boundaries on
the precise start and end time indices of every MoI.
Given such fine-grained labelling, models can generate MoIs from
the original videos to learn the best alignment of MoIs with their descriptions (Fig.~\ref{fig:idea} (a)).
To avoid the high annotation cost and subjective annotation bias%
\footnote{Different temporal boundaries are marked for the same sentences~\cite{anne2017didemo}.},
recent works focus on weakly-supervised learning without per-sentence
temporal boundary annotations in training~\cite{duan2018wsdec,gao2019wslln,mithun2019tga}.

Existing weakly-supervised solutions~\cite{zhang2020ccl,tan2020logan,lin2020scn} 
localise different MoIs \sgg{individually} (Fig.~\ref{fig:idea} (b)), 
which is not optimal as it neglects the fact that
the cross-sentence \sgg{relations in} a paragraph 
play an important role in temporally localising multiple MoIs. 
\sgg{Critically, an} individual sentence is sometimes ambiguous 
out of \sgg{its} paragraph context~\cite{yan2019leveraging,stolcke2000dialogue,zimmermann2009joint}. 
For example in Fig.~\ref{fig:idea} (c), 
without the consideration of the \textit{temporal relations} with the second sentence, 
the first query sentence (purple) can be easily mismatched with incorrect video segment, 
which is visually indiscriminate from the ground-truth moment.
Our analysis on the	
ActivityNet-Captions~\cite{krishna2017anet_captions} shows that	
the temporal relations of over $65\%$ moment pairs predicted by 
\raymond{a latest model~\cite{lin2020scn}}
are contradictory with the true order of their descriptions. 
\sgg{Yet}, MoIs described by \sgg{a paragraph} are often
semantically related to each other \sgg{in their corresponding
  sentences}. 
For example in Fig.~\ref{fig:idea} (d), ``the man'' in the blue query exhibits ambiguity 
if its \textit{semantic relations} with previous sentences are ignored. 
We also observed that 
more than $38\%$ descriptions in ActivityNet-Captions~\cite{krishna2017anet_captions} 
contain \sgg{ambiguous ways of} referring \sgg{to} expressions, \eg pronouns.
\sgg{To conclude, there are large error-margins in mis-localising individual
sentences to video segments in isolation.}

In this work,
we introduce a weakly-supervised method
for video activity localisation by natural language
called \textit{\method{full}} (\method{abbr}).
The key idea is to
explore the cross-sentence relations in \sgg{a} paragraph 
as \sgg{constraints} to \sgg{better interpret and match}
complex moment-wise temporal and semantic relations in videos.
Given the one-to-one moment-sentence mappings,
the inherent cross-moment relations
are unknown and not straightforward to be modelled in videos
but intrinsically available in the paragraph descriptions.
Hence, we impose the same cross-sentencing relations 
to their potentially matching video moments
for more reliable proposal selections.
The proposed \method{abbr} method differs significantly from 
the existing weakly-supervised models~\cite{zhang2020ccl,ma2020vlanet,tan2020logan}
which localise per-sentence queries \textit{\sgg{individually}}. 
They lack fundamentally any ability to make use of the cross-sentence relations
for moment proposal selection in model training.
Even though such relational information is less complete than		
per-sentence fine-grained temporal annotation, 		
it requires no annotation and avoids subjective bias from inherent		
ambiguity in temporal labelling~\cite{anne2017didemo}.	
Specifically,
by assuming different activities \sgg{in videos} are described sequentially,
we formulate a \textit{temporal consistency} constraint to encourage 
the selected moments to be temporally ordered according to their
descriptions in a paragraph (Fig.~\ref{fig:idea} (c)).
\raymond{
This is different from 
the temporal pretext tasks 
in self-supervised video learning
where the temporal constraint is adopted within a \textit{single modality}.
We exploit it in a \textit{cross modality} setup,
\ie, 
constraining the temporal order of event in visual modality
by the sentences order in text modality.}
%
Moreover, 
we encourage moment proposal selections to satisfy cross-sentence
\sgg{broader} semantics \sgg{in context}
to \sgg{minimise video-text matching} ambiguities.
To that end, we introduce a \textit{semantic consistency} constraint
to ensure that a moment selected for any pairing of two	
sentences (concatenation) in a paragraph is consistent (overlapping) with 	
the union of the selected segments per sentence (Fig.~\ref{fig:idea} (d)).

Our \sgg{{\bf contributions} are}:
\textbf{(1)}
To our best knowledge, 
this is the first idea to develop a model using \textit{cross-sentence relations} 
in a paragraph to 
explicitly represent and compute \textit{cross-moment relations} in videos,
so as to alleviate the ambiguity of \sgg{each individual} sentence \sgg{in} video activity localisation.
%
\textbf{(2)} We formulate a new weakly-supervised method 
for activity localisation by natural language 
called \textit{\method{full}} (\method{abbr}), that trains \sgg{a
model} with both temporal and semantic cross-sentence relations to improve
per-sentence temporal boundary prediction in testing.
\textbf{(3)}  Our approach achieves \sgg{the} state-of-the-art performance 
on two available activity localisation benchmarks,
especially so given more complex query descriptions.


\section{Related Works}
Early studies of video activity localisation by natural language mostly concentrate on 
making use of temporal annotations
to learn visual-text alignment with \textit{strong supervision}%
~\cite{gao2017tall,chen2018TGN,ghosh2019excl,zhang2019man,zeng2020drn,chen2020hvtg}.
However,
due to the unaffordable annotation cost of the fine-grained temporal boundary,
a growing number of works in recent years
have turned to tackle this task 
with only the video-level moment's description,
\ie \textit{weak supervision}~\cite{duan2018wsdec,gao2019wslln,mithun2019tga,lin2020scn,wu2020bar,zhang2020ccl}.

\paragraph{Strong Supervision.}
With the help of temporal annotation,
fully-supervised methods
localise activity in untrimmed videos
either in frame or segment-level.
SAP~\cite{chen2019semantic}
proposed to compute the visual-linguistic correlation scores
of the sentences and every frame in videos
and group the highly correlated frames as the predicted moments.
MCN~\cite{hendricks2018localizing} instead pre-divided videos into candidate segments (proposals)
with variant lengths in different positions
so to conduct segment-level semantic alignment.
The latest methods either follow SAP
to predict the probabilities of boundary across frames~\cite{Chen2019localizing,zeng2020drn,chen2020hvtg,chen2020rethinking,mun2020lgi}
or in the same spirit as MCN
to select from a set of pre-defined proposals
constructed by explicit sliding windows~\cite{gao2017tall,liu2018cross}
or implicit multi-granularity anchors~\cite{zhang2019man,xu2019multilevel,yuan2019semantic}.
Recently,
DPIN~\cite{wang2020dpin} 
proposed to combine the two localisation strategies
by a dual path interaction network
so to take the advantage of both.
Regardless of their remarkable success,
fully-supervised methods rely heavily on the fine-grained temporal annotation,
which is not only expensive but also prone to subjective bias~\cite{anne2017didemo}.
In this work,
we propose to further exploit the video-level descriptions of MoIs
as well as their relations
so to reduce the gaps between the weakly and fully-supervised models
without extra annotation cost.

\paragraph{Weak Supervision.}
In the absence of temporal boundary annotations,
most of the existing weakly-supervised approaches 
are either based on multi-instance learning~\cite{keeler1991mil} (MIL)
or jointly learn with reconstruction task.
The MIL-based methods~\cite{gao2019wslln,tan2020logan,ma2020vlanet,zhang2020ccl}
learn the visual-text alignment in the video-level by
maximising the matching scores of the videos and their corresponding descriptions 
manually annotated on the datasets
while suppressing that of the videos and the descriptions of others.
Such learned visual-text alignment is then applied to 
localise the moments which are best matched with the given queries
in inference.
Another commonly adopted strategy~\cite{lin2020scn,duan2018wsdec}
aims at selecting the video segments 
which can help accomplish the reconstruction task to the largest extent,
\eg WS-DEC~\cite{duan2018wsdec} jointly optimises the sentence localisation and video captioning tasks
so to identify the video segments which yield consistent captions with the queries.
Even though remarkable progress has been made in the past few years,
none of these methods fully exploit the video-level descriptions
but treat different sentences in the paragraph independently.
In this work,
we propose to explore the relations of sentences in paragraphs
to constrain the selections of moments in training
so that only the reliable video segments with consistent relations
will be aligned with the query sentences.

Temporal action localisation~\cite{gao2017CBR,lin2018bsn,zhao2017temporal}
is a similar task
which localises the pre-defined action classes in untrimmed videos.
However, the language query is usually composed of multiple actions with intricate correlations,
which make it more practical but challenging to be localised.

\section{Weakly-Supervised Activity Localisation}

Suppose we have $N$ untrimmed video $\mathcal{V} = \{V_i\}_{i=1}^N$
with each composed of $L_c$ disjoint clips $V_i = \{c_i^j\}_{j=1}^{L_c}$ in fixed duration.
Corresponding to each video,
we have a description paragraph consisting of 
$L_q$ text query sentences $Q_i = \{Q_i^j\}_{j=1}^{L_q}$ one-to-one describing the MoIs in $V_i$.
Given a video-query pair $(V_i, Q_i^j)$,
by dividing the untrimmed video $V_i$ into $L_s$ candidate segments $\{S_i^k\}_{k=1}^{L_s}$ 
using sliding windows~\cite{lin2020scn,ma2020vlanet} as the \textit{proposals},
our objective
is to select the $S_i^k$ from all the proposals 
which is most aligned with $Q_i^j$ \textit{in semantic}.
For simplicity, 
we take a single video $V$ and its description paragraph $Q = \{Q^j\}_{j=1}^{L_q}$ as example in the following discussion
and deprecate the subscript $i$.
Although the video-query (multi-sentences) relations are available in training, there is
no access to the ground-truth per-sentence temporal boundary. This is
a weakly-supervised learning problem where video proposals $S^k$
interact with the text queries $Q^j$ to discover the most plausible matches
between video segments and text sentences.

\begin{figure*}[ht]
\centering
\includegraphics[width=1.0\linewidth]{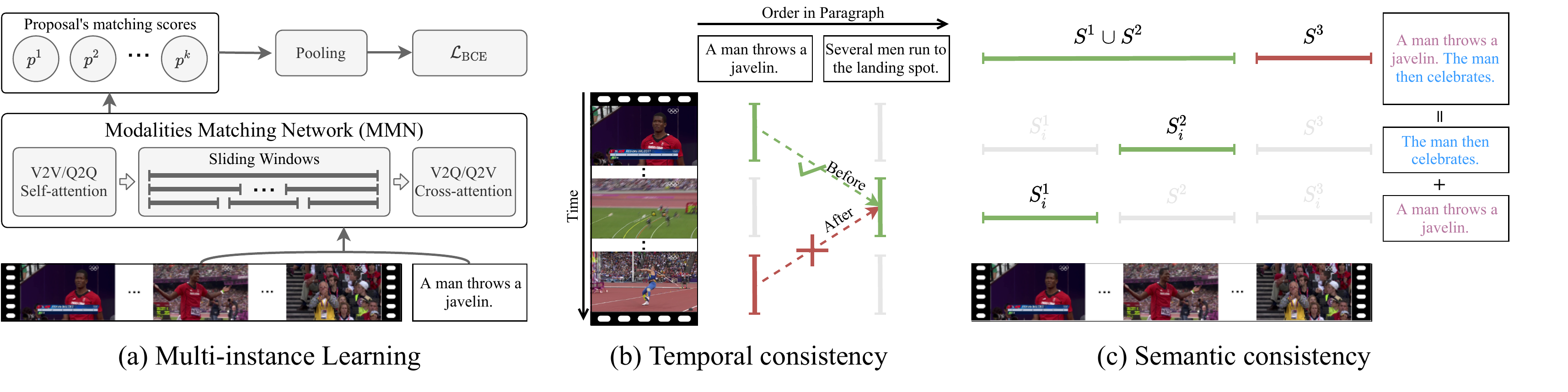}
\caption{
Overview of the proposed \textit{\protect\method{full}} (\protect\method{abbr}) method. 
\textbf{(a)} The modalities matching network (MMN)
is composed of self and cross-attention units 
and trained by MIL objective. 
\textbf{(b)}
The joint matching scores
of two queries to a pair of proposals
are optimised to encourage the consistency of cross-sentence and cross-moment temporal relations.
\textbf{(c)}
A longer query is synthesised by pairs of sentences in a paragraph (concatenation),	
whose their pairwise localisation is constrained to be consistent	
with the union of the two proposals selected for each sentence.
}
\label{fig:overview}
\end{figure*}

Here we formulate a \textit{\method{full}} (\method{abbr}) method
for this task. Fig.~\ref{fig:overview} shows an overview.
We first learn the visual-text alignment in video-level
with the same spirit of MIL
to feed a video-query pair into a \textit{modalities matching network} (MMN),
which predicts the matching score of the query and every proposal 
and supervise the max-pooling of scores by binary cross-entropy loss.
We then explore 
the order of two descriptions in the paragraph
and optimise their joint matching scores
to a proposals pair with consistent temporal relations.
\sgg{Furthermore}, we synthesise a longer query 	
by forming pairs of sentences in a paragraph (concatenation)	
and encourage its pairwise localisation to be semantically consistent 	
with the union of proposals individually selected for each sentence.	
\sgg{This is to minimise} the ambiguities in sentences
\sgg {so to improve} the model's interpretation of multiple video moments	
in a more complex \sgg{sentencing} context.

\subsection{Video-Sentence Alignment}

We start with the alignment of representations from two different modalities,
\ie an untrimmed video $V = \{c_i^1, c_i^2, \cdots, c_i^{L_c}\} \in \mathbb{R}^{L_c \times D_v}$ 
composed of $L_c$ clips
and a query sentence $Q^j = \{w^{j,1}, w^{j, 2}, \cdots, w^{j, L_w}\} \in \mathbb{R}^{L_w \times D_t}$
with $L_w$ words.
To explore the relation of $V$ and $Q^j$
and enable visual-text interaction,
both the representations are first projected into $D$-dimensional spaces 
by two independent fully-connected layers, respectively.
For clarity concern,
we reuse the symbols $V \in \mathbb{R}^{L_v \times D}$ and $Q^j \in \mathbb{R}^{L_w \times D}$
after projections.
Both the video $V$ and the query $Q^j$ will then be fed into a \textit{Modalities Matching Network} (MMN),
which will generate a set of candidate moments (proposals) $\{S^1, S^2, \cdots, S^{L_s}\}$ by sliding windows~\cite{lin2020scn,ma2020vlanet}
and predicts their individual matching scores 
with the input query $\{p(S^k \vert Q^j)\}_{k=1}^{L_s}$ (Fig.~\ref{fig:overview} (a)).
Motivated by the remarkable success of Transformer~\cite{vaswani2017transformer,devlin2018bert} on sequence analysis,
the MMN is composed of a stack of attention units to explore both the within and cross-modal correlation.

\paragraph{Attention Unit.}
As the building block of our MMN,
the attention unit plays a significant role
to learn the representation of a target sequence
in terms of its correlations with every element 
in a reference sequence.
Given a target sequence $X^t \in \mathbb{R}^{L_t \times D}$
and a reference  $X^r \in \mathbb{R}^{L_r \times D}$,
an attention unit $\mathcal{F}(X^t, X^r)$
attends $X^t$ using $X^r$ as follows:
\begin{equation}
\begin{gathered}
\mathcal{A} = \text{Softmax}(X^t{W^q}^\top W^k{X^r}^\top / \sqrt{D}) \in \mathbb{R}^{L_t \times L_r} \\
\mathcal{F}(X^t, X^r) = \text{FC}(X^t + \mathcal{A}X^r{W^v}^\top) \in \mathbb{R}^{L_t \times D}.
\end{gathered}
\label{eq:attention}
\end{equation}
The notions
\{$W^q; W^k; W^v\} \in \mathbb{R}^{3 \times D \times D}$ in Eq.~\eqref{eq:attention}
are three learnable matrices
and the coefficient $1 / \sqrt{D}$ is 
to counteract the effect of small gradients
caused by large $D$~\cite{vaswani2017transformer}.
The $\text{Softmax}(\cdot)$ is the row-wise softmax normalisation
and $\mathcal{A}$ is the correlation scores of target-reference element pairs.
The $\text{FC}(\cdot)$ is a linear projection with consistent input-output dimensions.
The attended result serves as the updated representation of the target sequence.

To investigate the visual-text matching relations,
it is essential to explore not only the within-modal context
but also the cross-modal interaction~\cite{ma2020vlanet}.
Hence,
the MMN is constructed by both self-attention and cross-attention blocks.
The video $V$ and the query $Q^j$
are first fed into two independent self-attention blocks respectively,
in which the target and reference inputs are from the same modalities:
\begin{equation}
V \leftarrow \mathcal{F}^\text{V2V}(V, V),\quad
Q^j \leftarrow \mathcal{F}^\text{Q2Q}(Q^j, Q^j).
\label{eq:self_attn}
\end{equation}
By doing so,
the salient clips/words in the input video/query are highlighted
by considering the context of the video or sentence. 
Conventional sliding window strategy~\cite{lin2020scn,ma2020vlanet} 
is then adopted to divide the video into $L_s$ proposals
$V = \{S^k\}_{k=1}^{L_s} \in \mathbb{R}^{L_s \times D}$.
Each proposal is composed of arbitrary continual clips in $V$
and represented by max-pooling the features of its included clips.
After that,
the two representations are interacted by cross-attention blocks:
\begin{equation}
V \leftarrow \mathcal{F}^\text{Q2V}(V, Q^j),\quad
Q^j \leftarrow \mathcal{F}^\text{V2Q}(Q^j, V),
\label{eq:cross_attn}
\end{equation}
which attends one modality by another
so to suppress the redundant text and irrelevant visual information.

\paragraph{Matching Score.}
Given the visual features $V = \{S^k\}_{k=1}^{L_s}$ 
and the text representation $Q^j = \{w^{j,k}\}_{k=1}^{L_w}$,
the matching score $p(S^k \vert Q^j)$ of a proposal-query pair 
is predicted according to both the modalities.
The sentence representation is first computed by aggregating all the words:
$Q^j \leftarrow \cmax(\{w^{j,k}\}_{k=1}^{L_w}) \in \mathbb{R}^{1 \times D}$
where $\cmax(\cdot)$ denotes the column-wise max-pooling function,
which is then fused with every proposal's representation~\cite{gao2017tall,hendricks2018localizing}:
\begin{equation}
E^{k,j} = (S^k + Q^j) \Vert (S^k \otimes Q^j) \Vert \text{FC}(S^k \Vert Q^j).
\label{eq:fusion}
\end{equation}
The notion $(\cdot \otimes \cdot)$ indicates the element-wise multiplication and 
$(\cdot \Vert \cdot)$ is the concatenation of two vectors
while $\text{FC}(\cdot)$ standing for a linear projection.
After that,
the joint representations $\{E^{k,j}\}_{k=1}^{L_s}$ are fed into a linear classifier:
\begin{equation}
p(S^k \vert Q^j) = \sigma(E^{k,j}W^\top + B).
\label{eq:score}
\end{equation}
The variable $\{W,B\}^\top \in \mathbb{R}^{D+1}$ is the weights of classifier 
and $\sigma(\cdot)$ is the sigmoid function.
The yielded probabilities $\{p(S^k \vert Q^j)\}_{k=1}^{L_s} \in (0, 1)$ 
serve as the matching scores
between proposals and query,
which is abbreviated to $p^{k,j}$.

\paragraph{Multi-Instance Learning.}
In the absence of temporal boundary,
the ground-truth moment is agnostic.
Therefore, 
we optimise the matching scores in video-level
to facilitate visual-text alignment.
To that end,
the matching score between the video $V$ and the query $Q^j$ is obtained by 
the max-pooling of all the proposals' score
$p(V\vert Q^j) \leftarrow \max(\{p^{k,j}\}_{k=1}^{L_s})$.
For each positive pair $(V, Q^j)$ given manually on the dataset,
we construct two negative counterparts by replacing either $V$ or $Q^j$
by a randomly sampled video $V^-$ or sentence $Q^-$ from the mini-batch
and compute their matching scores in the same way as $p(V \vert Q^j)$.
The binary cross-entropy (BCE) loss function is then adopted 
as the video-query alignment supervision signal:
\begin{equation}
\begin{aligned}
\mathcal{L}_\text{BCE}(V, Q^j) = 2 * &-\log p(V \vert Q^j) \\
-\log (1 - p(V \vert Q^-)) &-\log (1 - p(V^- \vert Q^j)),
\end{aligned}
\label{eq:bce}
\end{equation}
where the coefficient $2$ is applied to the positive term considering the balance of positive and negative pairs.
The rationale behind Eq.~\eqref{eq:bce} is 
assuming that the MoIs in one video doesn't exist in any other videos
so $(V, Q^-)$ and $(V^-, Q^j)$ should be semantically unmatched.
By minimising $p(V \vert Q^-)$ and $p(V^- \vert Q)$,
the predictions of the incorrect proposals in $V$ with different semantics from $Q^j$ 
will also be minimised implicitly
so that the learned matching scores
can reveal the inherent visual-text relations.
This takes the spirit of MIL~\cite{keeler1991mil} by treating the proposals as the instances in a bag (video)
and learning with the bag-level annotations.

\subsection{Cross-Sentence Relations Mining}

The $\mathcal{L}_\text{BCE}$ in Eq.~\eqref{eq:bce} 
aligns queries with the proposals
yielding the largest matching scores among all the candidates. 
However, the predicted scores can be unreliable 
due to the visually indiscriminate moment proposals existed in videos
and text ambiguities in \sgg{individual sentences}
which will lead to visual-text \sgg{misalignment in training}.
Therefore,
we explore the cross-sentence relations
to select reliable proposals with consistent cross-moment relations.

\paragraph{Temporal Consistency.}
As the video frames are naturally exhibited to the viewers in time order,
the temporal relations of different MoIs should intrinsically be encoded in 
the order of their descriptions in the paragraph.
With such an assumption,
we can identify
the pairs of proposals 
both yielding high predicted matching score 
with the corresponding queries
but inconsistent in temporal relations,
which are likely to be incorrect.
Given arbitrary query sentences pair $(Q^j, Q^{j'})$ from the description paragraph of video $V$,
their respective selected segments $(S^k, S^{k'})$ should satisfy similar temporal structure with them,
\ie $S^k$ should occur before $S^{k'}$ in the video
if $Q^j$ is in front of $Q^{j'}$ in the paragraph and vice versa.
The temporal order of two proposals
$\mathcal{R}(S^k, S^{k'})=0$ if $S^k$ starts before $S^{k'}$ in the video,
otherwise $\mathcal{R}(S^k, S^{k'})=1$.
Similarly, 
$\mathcal{R}(Q^j, Q^{j'}) = \mathbbm{1}[j >= j']$ 
where $j$ and $j'$ are the position of sentences
in the paragraph.
The temporal constraint is then formulated to ensure
$\mathcal{R}(S^k, S^{k'})=\mathcal{R}(Q^j, Q^{j'})$.

By assuming the matching scores of different queries to any proposals are independent,
the joint probability of $Q^j$ and $Q^{j'}$ are respectively matching with $S^k$ and $S^{k'}$ is:
\begin{equation}
p(S^k, S^{k'} \vert Q^j, Q^{j'}) = p(S^k \vert Q^j) \cdot p(S^{k'} \vert Q^{j'}).
\label{eq:joint_scores}
\end{equation}
As shown in Fig.~\ref{fig:overview} (b),
we take the queries' order as the ground-truth 
for the temporal relation of the proposal pair.
\raymond{Given $Q^j$ and $Q^{j'}$,}
the joint probabilities set 
$\{p(S^k, S^{k'} \vert Q^j, Q^{j'})\}_{k,k'=1}^{L_s}$ 
is then divided into two subsets:
for all the proposal pairs $(S^k, S^{k'})$,
the joint probability $p(S^k, S^{k'} \vert Q^j, Q^{j'}) \in P^+_t$ if $\mathcal{R}(S^k, S^{k'}) = \mathcal{R}(Q^j, Q^{j'})$,
otherwise belonging to $P^-_t$.
The MIL loss is re-formulated with the temporal constraint:
\begin{equation}
\begin{aligned}
\mathcal{L}_\text{TMP}(V, Q^j, Q^{j'}) = 
&-\log (\max(P^+_t)) \\
&- \log (1 - \max(P^-_t)).
\end{aligned}
\label{eq:temporal}
\end{equation}
By training with $\mathcal{L}_\text{TMP}$,
the model learns to align the proposals with queries
only if they are temporally consistent.
This refrains the model from visual-text misalignment
in the absence of ground-truth temporal annotations.

\paragraph{Semantic Consistency.}
To \sgg{minimise} the negative \sgg{impact} from ambiguous
\sgg{per-sentence expressions in isolation}
and \sgg{to explore the context of a paragraph,
it is beneficial for a model to consider broader semantics 
beyond individual sentences by relating other expressed
objects/actions in a wider context}~\cite{mun2020lgi}.
However,
it is nontrivial to explicitly do so
since the object/action's information
is missing without fine-grained annotation.
In this case,
we propose to form pairs of MoIs by concatenation in the same videos:
$Q^{j,j'} = Q^j \Vert Q^{j'}$
and train the model to localise the concatenated longer query
with the consideration of both sentences in each pair.
Given the proposals $S^k$ and $S^{k'}$ with the largest
$p(S^k, S^{k'} \vert Q^j, Q^{j'})$ in Eq.~\eqref{eq:temporal},
the matching scores of $Q^{j,j'}$ and the video segments $S^l$ 
is optimised to encourage the consistency of $S^l$ and $S^k \cup S^{k'}$
(Fig.~\ref{fig:overview} (c)).
As in the temporal constraint,
we divide the predicted scores $p(S^l \vert Q^{j,j'})$ into two subsets:
for all the proposals $S^l$ in the video $V$,
$p(S^l \vert Q^{j,j'}) \in P^-_s$ if $\text{IoU}(S^l, S^k \cup S^{k'}) < \tau$,
and $P^+_s$ is composed of the $S^l$ which is most consistent with $S^k \cup S^{k'}$.
The $\tau$ decides 
how two proposals are deemed inconsistent 
regarding their intersection over union score (IoU)
which is set to $0.5$ in practice.
The constraint on the semantic consistency of $S^l$ and $S^k \cup S^{k'}$ is formulated as:
\begin{equation}
\begin{aligned}
\mathcal{L}_\text{SMT}(V, Q^j, Q^{j'}) = 
&-\log(\max(P^+_s)) \\
&-\log (1 - \max(P^-_s)).
\end{aligned}
\label{eq:semantic}
\end{equation}
To minimise $\mathcal{L}_\text{SMT}$,
the model is explicitly trained to consider the semantics of both $Q^j$ and $Q^{j'}$
when localising $Q^{j,j'}$
so to ensure the overlap of $S^l$ and $S^k \cup S^{k'}$.
\sgg{By} introducing additional \sgg{longer} queries
synthesised \sgg{from} pairwise sentences 
\sgg{in} model training,
\sgg{it enhances} the model's capacity \sgg{to interpret and match more complex
descriptions to video moments}, \sgg{critical in practice due to
that untrimmed raw videos are often unstructured}.

\subsection{Model Training}
In each training iteration,
we randomly sample $n$ videos with a pair of queries for each from its paragraph description 
as a mini-batch
and the overall loss is computed by:
\begin{equation}
\begin{aligned}
\mathcal{L} =& \frac{1}{2 * n}\sum_{i=1}^n\sum_{j=1}^2 \mathcal{L}_\text{BCE}(V_i, Q_i^j) \\
&+ \frac{1}{n}\sum_{i=1}^n \mathcal{L}_\text{TMP}(V_i, Q_i^1, Q_i^2) \\
&+ \frac{1}{n}\sum_{i=1}^n \mathcal{L}_\text{SMT}(V_i, Q_i^1, Q_i^2).
\end{aligned}
\label{eq:loss}
\end{equation}
Since the objective function $\mathcal{L}$ in Eq.~\eqref{eq:loss} is differentiable,
conventional stochastic gradient descent algorithm 
is adopted for end-to-end model training.
The overall process of a training iteration is summarised in Alg.~\ref{alg:whole}.
\begin{algorithm}[ht]
    \caption{Video activity localisation by \protect\method{abbr}} 
    \label{alg:whole}
    \textbf{Input:} 
    Untrimmed videos $\mathcal{V}$,
    Paragraph descriptions $\mathcal{Q}$. \\
    \textbf{Output:}
    An updated video activity localisation model. \\
    Sampling a random mini-batch of videos; \\
    Sampling two queries for each video from its paragraph; \\
    \textbf{foreach} video-query pair \textbf{do} \\
        \hphantom{~~} 
        Mapping video and query to $D$-dimensional spaces; \\
        \hphantom{~~} 
        Conducting V2V and Q2Q self-attention (Eq.~\eqref{eq:self_attn}); \\
        \hphantom{~~} 
        Generating proposals by sliding windows; \\
        \hphantom{~~} 
        Conducting V2Q and Q2V cross-attention (Eq.~\eqref{eq:cross_attn}); \\
        \hphantom{~~} 
        Fusing each proposal's feature with the query (Eq.~\eqref{eq:fusion}); \\
        \hphantom{~~} 
        Computing the proposal-query matching scores (Eq.~\eqref{eq:score}); \\
    \textbf{end foreach} \\
    Computing the objective loss (Eq.~\eqref{eq:loss}); \\
    Updating model weights by back-propagation.
\end{algorithm}

\section{Experiment}

\paragraph{Datasets.}
Experiments were conducted on two video activity localisation datasets:
(1) Charades-STA~\cite{gao2017tall} 
contains 12,408/3720 video-query pairs from 5338/1334 videos for training and testing, respectively.
The query sentences are composed of 7.2 words on average
and the average duration of the target video moments and untrimmed videos are 8.1 and 30.6 seconds;
(2) ActivityNet-Captions~\cite{krishna2017anet_captions} is a much larger-scale dataset composed of 
19,290 videos with 37,417/17,505/17,031 MoIs in the train/val\_1/val\_2 split.
The average length of queries is 14 words
while that of the MoIs and untrimmed videos are 36.2 and 117.6 seconds.

The activities captured in those two datasets 
are of various complexity:
Only 6\% of the descriptions involve more than one actions in Charades
whilst 44\% in ActivityNet 
with 12\% \vs 44\%
regarding the number of people~\cite{lei2020tvr}.

\paragraph{Performance Metric.}
We followed previous works~\cite{duan2018wsdec,wu2020bar,chen2020hvtg}
to evaluate the activity localisation results
by the ``IoU@$m$'' metric where $m$ is 
the pre-defined temporal Intersection over Union (IoU) thresholds.
Given the temporal boundary $(s, e)$ of a target moment
and the selected segment proposal $(\tilde{s}, \tilde{e})$ 
with the largest predicted matching score, 
the IoU between the two video segments is computed by
$\frac{\max(0, \min(e, \tilde{e}) - \max(s, \tilde{s}))}{\max(e, \tilde{e}) - \min(s, \tilde{s})}$.
A prediction is considered correct if its IoU with the ground-truth is greater
than the pre-defined IoU thresholds set to $\{0.1, 0.3,
0.5\}$ on ActivityNet and $\{0.3, 0.5, 0.7\}$ on
Charades~\cite{duan2018wsdec,wu2020bar}.

\paragraph{Implementation.}
\raymond{
We used VGG (4096-D) and ResNet152 (2048-D) feature representations 
officially released with the datasets
for per-frame representations in Charades and ActivityNet, respectively.
The videos were truncated evenly (and zero-padded)
into 128 clips in Charades and 256 in ActivityNet,
with each clip represented by 
the max-pooling of 5 continual frame's features.}
The pre-trained GloVe embedding~\cite{pennington2014glove} was adopted
as the word feature representation (300-D)
and the maximal sentence length was set to 20 words.
Both the clip and word representations were 
linearly mapped to 256-D spaces
before being fed into MMN.
The sliding windows stride was $8$ and the window sizes were
$\{8, 12, 20, 32, 64\}$ in Charades and $\{8, 16, 32, 64, 128\}$ in ActivityNet.
\raymond{
The temporal dependencies
of video segments in terms of the same query sentences
were explored by an additional self-attention unit
before predicting their matching scores.}
As the paragraph descriptions were pre-divided into individual sentences on both datasets,
we restored the order of sentences in the paragraph
by the ground-truth start time of MoIs.
Note that timestamps were unavailable in proposal selections, neither in training nor testing.
The proposed \method{abbr} was trained 50 epochs by Adam optimiser
with a batch size of 64 and learning rate of $1e-4$.
Cross-sentence relations were only used in
training with no extra computational cost in testing.

\subsection{Comparisons to the State-Of-The-Art}
\begin{table}
\begin{subtable}[c]{1.\columnwidth}
\footnotesize\setlength{\tabcolsep}{0.1cm}
\begin{tabular}{|l|l|c|c|c|c|c|}
\hline
Split & Method & Moment & Query & 
IoU@0.1 & IoU@0.3 & IoU@0.5 \\
\hline\hline
\multirow{5}{*}{val\_2} 
& DPIN~\cite{wang2020dpin} & \cmark & \xmark 
& - & 62.40 & 47.27 \\
& 2D-TAN~\cite{zhang2020tan} & \cmark & \xmark 
& - & 59.45 & 44.51 \\
& DRN~\cite{zeng2020drn} & \xmark & \xmark 
& - & -  & 42.49 \\
& LGI~\cite{mun2020lgi} & \cmark & \xmark 
& - & 58.52 & 41.51 \\
& HVTG~\cite{chen2020hvtg} & \xmark & \xmark 
& - & 57.60 & 40.15 \\
\hline\hline
\multirow{4}{*}{val\_1} 
& WS-DEC~\cite{duan2018wsdec} & \xmark & \xmark 
& 62.71 & 41.98 & 23.34 \\
& WSLLN~\cite{gao2019wslln} & \xmark & \xmark 
& \second{75.4} & 42.8 & 22.7 \\
& BAR~\cite{wu2020bar} & \cmark & \xmark 
& - & \second{49.03} & \second{30.73} \\
\cline{2-7}
& \textbf{\protect\method{abbr}} (Ours) & \cmark & \cmark 
& \best{76.66} & \best{51.17} & \best{31.67} \\
\hline
\multirow{4}{*}{val\_2} 
& SCN~\cite{lin2020scn} & \cmark & \xmark 
& 71.48 & 47.23 & 29.22 \\
& RTBPN~\cite{zhang2020rtbpn} & \cmark & \xmark 
& \second{73.73} & 49.77 & 29.63 \\
& CCL~\cite{zhang2020ccl} & \cmark & \xmark 
& - & \second{50.12} & \second{31.07} \\
\cline{2-7}
& \textbf{\protect\method{abbr}} (Ours) & \cmark & \cmark 
& \best{81.61} & \best{55.26} & \best{32.19} \\
\hline
\multirow{2}{*}{OOD} 
& WS-DEC~\cite{duan2018wsdec} & \cmark & \xmark 
& 30.71 & 17.00 & 7.17 \\
\cline{2-7}
& \textbf{\protect\method{abbr}} (Ours) & \cmark & \cmark 
& \best{38.35} & \best{22.77} & \best{10.31} \\
\hline
\end{tabular}
\caption{ActivityNet-Captions}
\end{subtable}
\\
\begin{subtable}[c]{1.\columnwidth}
\small\setlength{\tabcolsep}{0.1cm}
\begin{tabular}{|l|c|c|c|c|c|}
\hline
Method & Moment & Query & 
IoU@0.3 & IoU@0.5 & IoU@0.7 \\
\hline\hline
DPIN~\cite{wang2020dpin} & \cmark & \xmark 
& - & 47.98 & 26.96 \\
2D-TAN~\cite{zhang2020tan} & \cmark & \xmark 
& - & 39.81 & 23.25 \\
DRN~\cite{zeng2020drn} & \xmark & \xmark 
& - & 53.09 & 31.75 \\
LGI~\cite{mun2020lgi} & \cmark & \xmark 
& 72.96 & 59.46 & 35.48 \\
HVTG~\cite{chen2020hvtg} & \xmark & \xmark 
& 61.37 & 47.27 & 23.30 \\
\hline\hline
TGA~\cite{mithun2019tga} & \xmark & \xmark 
& 29.68 & 17.04 & 6.93 \\
SCN~\cite{lin2020scn} & \cmark & \xmark 
& 42.96 & 23.58 & 9.97 \\
LoGAN~\cite{tan2020logan} & \cmark & \xmark 
& 51.67 & \second{34.68} & 14.54 \\
BAR~\cite{wu2020bar} & \cmark & \xmark 
& 44.97 & 27.04 & 12.23 \\
RTBPN~\cite{zhang2020rtbpn} & \cmark & \xmark 
& \best{60.04} & 32.36 & 13.24 \\
VLANet~\cite{ma2020vlanet} & \cmark & \xmark 
& 45.24 & 31.83 & 14.17 \\
CCL~\cite{zhang2020ccl} & \cmark & \xmark 
& - & 33.21 & \second{15.68} \\
\hline
\textbf{\protect\method{abbr}} (Ours) & \cmark & \cmark
& \second{53.66} & \best{34.76} & \best{16.37} \\
\hline
\end{tabular}
\caption{Charades-STA}
\end{subtable}
\caption{Performance comparisons on video activity localisation methods. 
\raymond{Fully and weakly-supervised methods are shown 
in the upper and lower part of each table, respectively.}
The `Moment' column refers to methods trained by
exploiting multiple video moments corresponding to the same-sentence,
whilst the `Query' column refers to training by cross-sentence temporal
ordering and sentence pairing in the context of a paragraph.
The `Split' column denotes the different data splits in the ActivityNet-Captions 	
used in the evaluations.
\raymond{The discounted recall rates~\cite{yuan2021closer} are reported
for the `OOD' split of ActivityNet-Captions.}
}
\label{tab:cmp}
\vspace{-0.5cm}
\end{table}
Table~\ref{tab:cmp} compares the performance of \method{abbr}
against the state-of-the-art video activity localisation models
including both fully- and weakly-supervised methods.
We observe:
\textbf{(1)} Not surprisingly, fully-supervised models outperform
weakly-supervised models clearly. 
However, CRM reduces that
performance gap by over 41\%
on the ActivityNet at
$\text{IoU}=0.3$.
\textbf{(2)} Discovering different video moments 
correlating to the \textit{same-sentence} for proposal selection
has been exploited to a good effect by existing methods in the form of
attention~\cite{lin2020scn,ma2020vlanet} or 2D temporal
convolution~\cite{zhang2020rtbpn,zhang2020tan}.
However, the notably better performance of \method{abbr} compared to
those methods further demonstrates the additional advantage of 
using {\em cross-sentence} temporal and semantic relations within a paragraph
for learning better visual-text alignment
and benefiting per-sentence localisation in testing.
\textbf{(3)} \method{abbr} surpasses the state-of-the-art
weakly-supervised methods 
across the board
except for IoU@0.3 on Charades.
This demonstrates compellingly the effectiveness of \method{abbr} from
modelling explicitly cross-sentence relations.
\raymond{
Our advantages
on the OOD split of
ActivityNet-Captions~\cite{yuan2021closer}
further indicate CRM's better multi-modal understanding
rather than driven by annotation biases.}

\subsection{Components Analysis}
We investigated the effects of different components in CRM model design to 
\raymond{
study their individual contributions.
The ``val\_1'' split of ActivityNet was adopted.
}

%
\begin{figure}[ht]
\centering
\includegraphics[width=1.0\linewidth]{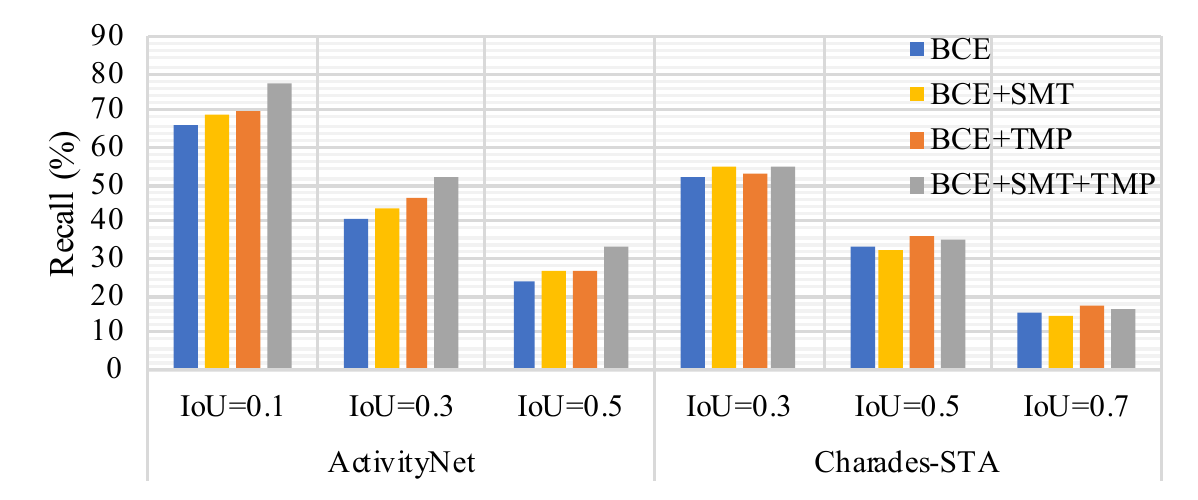}
\caption{
Effects of cross-sentence relations mining. 
BCE is the base model trains with only the MIL objective (Eq.~\eqref{eq:bce}).
TMP and SMT are the proposed constraints on 
temporal (Eq.~\eqref{eq:temporal}) and semantic (Eq.~\eqref{eq:semantic}) relational consistency.
}
\label{fig:loss}
\end{figure}
\paragraph{Effects of Cross-sentence Relations.}
We evaluated the effectiveness of 
imposing cross-sentence relational consistency
by training the baseline model (BCE) with either the temporal (BCE+TMP) or semantic (BCE+SMT) constraint
as well as with both (BCE+TMP+SMT).
Fig.~\ref{fig:loss} shows that both constraints are
beneficial individually 
and the benefits become more clear when they are jointly adopted.
Moreover, the performance improvement is more significant on ActivityNet than Charades.
Given the generally more complex activities in ActivityNet, 
this shows that training CRM on combinations of pairwise
sentencing as semantic consistency constraint (Eq.~\eqref{eq:semantic})
has its unique advantages in activity localisation against more
complex query descriptions.

\paragraph{Temporal Consistency.}
\begin{table}
\setlength{\tabcolsep}{0.1cm}
\begin{center}
\begin{tabular}{|c|c|c||c|c|}
\hline
& \multicolumn{2}{|c||}{Train} & \multicolumn{2}{c|}{Test} \\\hline
Temporal & ActivityNet & Charades & ActivityNet & Charades \\\hline\hline
\xmark & 64.28 & 73.88 & 45.02 & 73.91 \\\hline
\cmark & 82.43 & 74.88 & 70.82 & 74.65 \\
\hline
\end{tabular}
\end{center}
\caption{Temporal consistency between the descriptions of MoI pairs and their selected proposals.
Metric: accuracy.}
\label{tab:temporal}
\vspace{-0.1cm}
\end{table}
%
\raymond{To verify our assumption on temporal order,
we compared how many correct predictions learned with and without
$\mathcal{L}_\text{TMP}$ (Eq.~\eqref{eq:temporal}) against the ground-truth.}
Specifically,
for each video consists of $n$ MoIs,
we constructed $C_n^2$ MoI pairs 
and measured the ratio of consistent pairs
\raymond{
by comparing the order of the two ground-truth moments 
and that of the selected proposals.}
Table~\ref{tab:temporal} shows that
by explicitly training CRM with cross-sentence temporal order constraint,
the video segments selected by \method{abbr} is much more consistent
in temporal relations on ActivityNet than the base models without it.
Although different moments in the test set are localised independently,
such advantages are still clear.
Besides, 
it is surprising to see that the cross-moment temporal relations
yielded by the base model on Charades 
are reasonably consistent with the 
\raymond{true order}
but the temporal constraint 
still benefited
the localisation results.
This implies the potential advantages of 
optimising joint matching scores of moment pairs with their descriptions
in learning effective visual-text alignment.

\paragraph{Semantic Consistency.}
\begin{table}
\setlength{\tabcolsep}{0.1cm}
\begin{center}
\begin{tabular}{|c|c|c||c|c|}
\hline
& \multicolumn{2}{|c||}{Train} & \multicolumn{2}{c|}{Test} \\\hline
Semantic & ActivityNet & Charades & ActivityNet & Charades \\\hline\hline
\xmark & 55.76 & 35.34 & 57.84 & 31.01 \\\hline
\cmark & 68.14 & 55.46 & 71.30 & 51.33 \\
\hline
\end{tabular}
\end{center}
\caption{Semantic consistency between the union of two MoIs' segments
and the one selected for the concatenation of their descriptions. 
Metric: prediction recall at $\text{IoU}=0.5$.}
\label{tab:semantic}
\vspace{-0.5cm}
\end{table}
As in the analysis of temporal consistency,
we enumerated all the possible MoI pairs in the same videos
and quantify the semantic consistency
by taking the union of MoI pairs
as the ground-truth moment corresponding to the concatenation of their descriptions.
More specifically,
given the sentence description of two MoIs and their temporal boundary $S^i$ and $S^j$,
we concatenated the two per-sentence queries and 
identified the video segment $S^k$ 
yielding the largest matching scores with the concatenation.
We then computed the temporal IoU between $S^i \cup S^j$ and $S^k$, where
$S^k$ is deemed semantically consistent with $S^i \cup S^j$ 
if $\text{IoU}(S^i \cup S^j, S^l) > 0.5$. 
\raymond{
Note that
it is not necessary for the two moments
to be consecutive in time
so that our semantic assumption can hold,
as the boundary defined by the concatenated description
always matches their temporal union.}
Table~\ref{tab:semantic} shows that
the baseline model trained without semantic constraint in Eq.~\eqref{eq:semantic}
yields sensible performances in localising the paired queries.
This demonstrates that CRM implicitly learns to consider the semantic context of queries
by the attention units. 
The superior results of CRM trained with explicit 
semantic constraint shows that it encourages broader consensus in
semantics across sentences. This explains why the performance advantages of \method{abbr}
is more significant when localising more complex activities
in ActivityNet.

%
\begin{figure}[ht]
\centering
\includegraphics[width=1.0\linewidth]{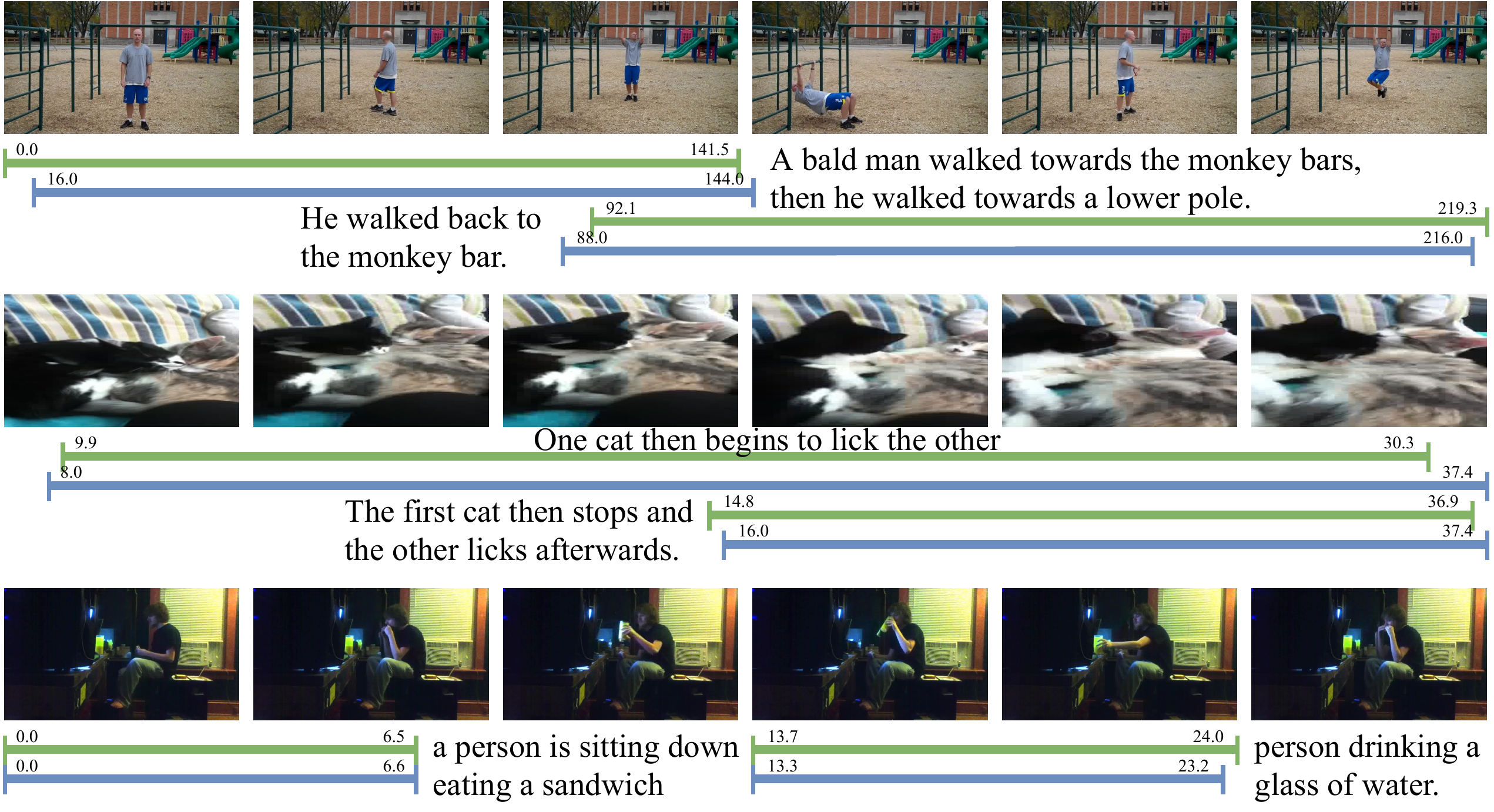}
\caption{
Qualitative examples show the interaction between MoIs in the same videos.
The green bars indicate the ground-truth MoI's boundaries
whilst the blue bars show the model predictions by \protect\method{abbr}.
The query sentences are simplified for illustrations only given the
space limit. 
}
\label{fig:cases}
\vspace{-0.1cm}
\end{figure}
\paragraph{Qualitative Examples.}
Fig.~\ref{fig:cases} shows some qualitative examples from both
ActivityNet and Charades. They show how different MoIs in the same videos
may interact with each other so that their relations can be used to optimise
per-sentence activity localisation in the context of a paragraph.
It is evident that
localising video moments by per-sentence independently 
is unreliable,
\eg in the first example (top-row), the man reaches the monkey bars 
both before and after he walks toward the lower pole.
``The first cat'' example in the middle-row is ambiguous without context.
By explicitly exploring the cross-sentence relations, CRM avoids such ambiguities
and minimises video-text misalignment. 

\begin{figure}[ht]
\centering
\begin{subfigure}[b]{1.\columnwidth}
\includegraphics[width=1.\columnwidth]{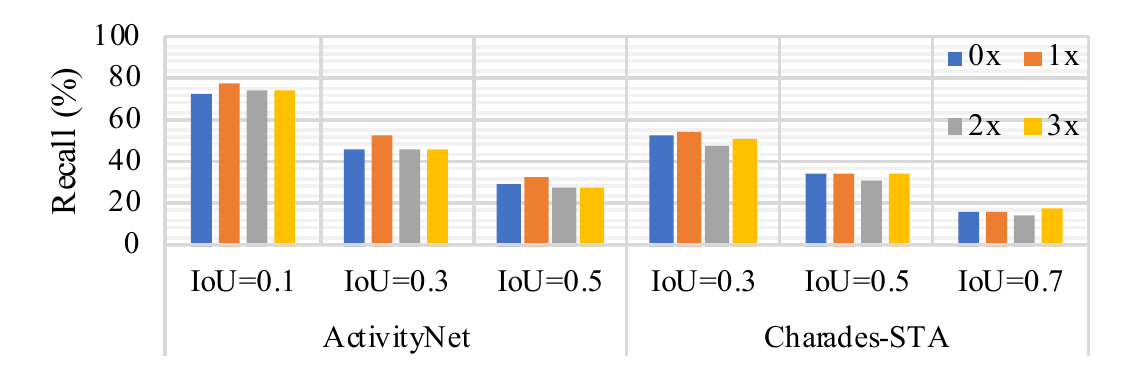}
\caption{Self-attention}
\label{subfig:self_attn}
\end{subfigure} 
\\
\begin{subfigure}[b]{1.\columnwidth}
\includegraphics[width=1.\columnwidth]{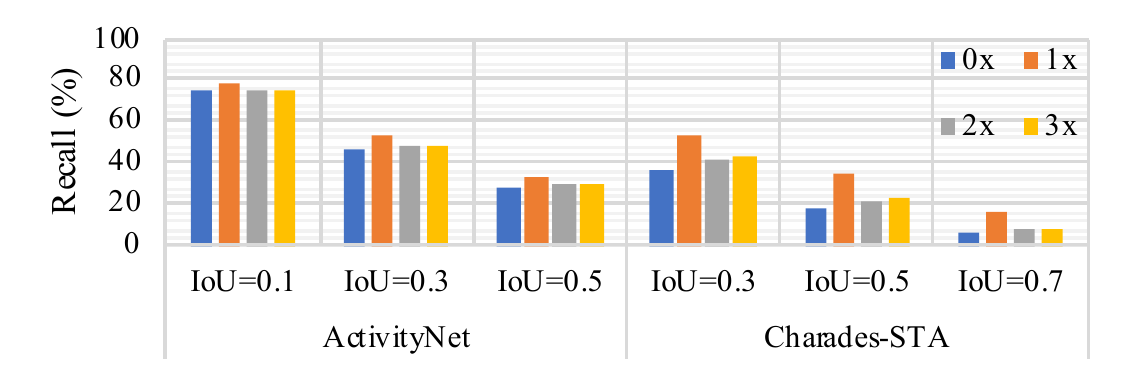}
\caption{Cross-attention}
\label{subfig:cross_attn}
\end{subfigure} 
\caption{Effects of attention units.
Models are constructed and trained with different numbers of self-attention and cross-attention units
to investigate their effects. 
}
\label{fig:attn}
\vspace{-0.3cm}
\end{figure}
\paragraph{Effects of Attention Units.}
As the building block of our MMN
backbone in section 3.1,
the attention units play a significant role 
in exploring the videos and sentences data
as well as their correlations.
We investigated its effect by comparing the prediction recall
of \method{abbr} constructed with different numbers of attention units,
showing its benefits in sequence analysis and visual-text interactions (Fig.~\ref{fig:attn}).
On the other hand, due to the limited video data available for training (10K/5K on ActivityNet/Charades),
stacking up attention layers
fails to further benefit CRM, leading to model performance degradation possibly due to overfitting.


\section{Conclusion}

In this work,
we presented a novel \textit{\method{full}} (\method{abbr}) method
for learning video activity localisation
in the absence of per-sentence temporal annotation. CRM
explores cross-sentence relations within each paragraph
description of a long video to optimise video moment proposal selections in training
so to improve per-sentence localisation in testing.
\method{abbr} minimises mis-matching individual sentences to video moment proposals during training
by constraining their selections according to 
the temporal ordering
and pairwise sentencing as expanded queries in the context of a paragraph description of video. 
This improves notably CRM's capacity to localise more accurately video
activities against more complex language descriptions.
%
Experiments on two available activity localisation benchmark datasets
show the performance advantages of the proposed \method{abbr} method
over a wide range of state-of-the-art weakly-supervised models.
Extensive ablation studies further provided
in-depth analysis of the effectiveness of the individual components in \method{abbr}.


\section*{Acknowledgements}
This work was supported by the China Scholarship Council, Vision Semantics Limited, the Alan Turing Institute Turing Fellowship, and Adobe Research.


{\small
\bibliographystyle{ieee_fullname}
\bibliography{press,egbib}
}

\end{document}